\def\year{2020}\relax
\documentclass[letterpaper]{article} 
\usepackage{aaai20}  
\usepackage{times}  
\usepackage{helvet} 
\usepackage{courier}  
\usepackage[hyphens]{url}  
\usepackage{graphicx} 
\usepackage{graphicx}  
\title{Integrating Intrinsic and Extrinsic Explainability: The Relevance of Understanding Neural Networks for Human-Robot Interaction}
\author{
{Tom Weber, Stefan Wermter}\\ 
Knowledge Technology, Department of Informatics, University of Hamburg\\
Vogt-Koelln-Strasse 30, 22527 Hamburg\\
\{tomweber, wermter\}@informatik.uni-hamburg.de 
}
 
\begin{document}

\maketitle

\begin{abstract}
Explainable artificial intelligence (XAI) can help foster trust in and acceptance of intelligent and autonomous systems. Moreover, understanding the motivation for an agent's behavior results in better and more successful collaborations between robots and humans. However, not only can humans benefit from a robot's explanation but the robot itself can also benefit from explanations given to him.
Currently, most attention is paid to explaining deep neural networks and black-box models. However, a lot of these approaches are not applicable to humanoid robots. Therefore, in this position paper, current problems with adapting XAI methods to explainable neurorobotics are described. Furthermore, NICO, an open-source humanoid robot platform, is introduced and how the interaction of intrinsic explanations by the robot itself and extrinsic explanations provided by the environment enable efficient robotic behavior.
\end{abstract}

\section{Introduction}
\label{sec:intro}
Neuro-inspired robots make use of connectionist models to tackle problems that are not easily solvable with symbolic methods and the success of complex deep neural networks (DNN) over the past decade made it possible for agents to acquire robust behavior through learning. Such flexible adaptation to problems, for example learning to distinguish objects just by touching them \cite{kerzel2019touch}, is still impossible to achieve with symbolic representations.
However, this advance in performance thanks to DNNs has disadvantages regarding explainable artificial intelligence since deep learning models can essentially be considered as black-box models to the human interpreter. The sub-symbolic nature of their distributed representations is not inherently explainable, i.e. incomprehensible to a human observer and needs post hoc explanations. In contrast, symbolic approaches like first-order logic or planners are inherently explainable but largely unable to deal with certain problems that are crucial for effective robotic intelligence like the efficient perception and robust learning of complex environments.

Explanations that can be produced from the inner states and algorithms of an agent, i.e. that can be produced from within the agent, we describe as \textit{intrinsic explanations}. For instance, explaining a classification decision which is done by a perceptive module, with a description of its relevant features. Almost all existing literature on XAI falls within this category.
An autonomous robot, however, is situated in an environment. Through interaction with it, the robot can obtain information about prior uncertainties or knowledge gaps. Therefore, explanations that the robot receives from an external source in its environment are \textit{extrinsic explanations}.

Currently, explainable human-robot interaction (HRI) faces two major obstacles. Neural networks constitute an irreplaceable part of modern robotics. However, the development of successful explanation methods of deep architectures is still in its infancy, and the intrinsic explanations that an agent can deliver are respectively limited, especially regarding the constraints of HRI contexts. The generation of an explanation is not sufficient if the communication to the receiver fails due to unintelligible information representation or a lack of information. The low-level explanations though feature visualization provided by a lot of XAI methods rarely enable the end-user to interpret the agent's behavior unambiguously. Therefore, we will review a few promising techniques and their limitations for explainable robots.
Additionally, the communication between robots and humans narrows the choice of modalities that can be used to explain which renders numerous explainability methods unusable without modification.

Neural networks offer robots the possibility to adapt and learn from extrinsic explanations that they receive from the environment, in an interactive way. The robot can learn how to shape its intrinsic explanations to make them more suitable for its audience. XAI methods for HRI should be extended to an interactive framework between intrinsic and extrinsic explanations to fully account for the nature of such situations.

\section{XAI for Neurorobotics}
\label{sec:xai}
Methods for explaining data-driven models are largely treated separately from methods for explaining goal-driven agents (for reviews of the former, cf. \cite{arrieta2020explainable}, for the latter, cf. \cite{anjomshoae2019explainable}).

State-of-the-art neuro-cognitive humanoid robots, like the NICO (see figure \ref{fig:nico}), are best realized in a multimodal neuro-symbolic hybrid fashion \cite{wermter2017cml}. Symbolic processes represent high-level cognitive concepts like beliefs, goals, and intentions, which are grounded in and make use of neural architectures for complex cognitive processes like perception. Still, despite this interconnection, the explanation methods are conceptionally kept apart from each other, instead of being integrated. 
The PeCoX framework separates the explanation of hybrid agents into two submodules, the cognitive XAI for symbolic and the perceptive XAI for sub-symbolic processes \cite{neerincx2018pecox}. Both are detached from one another, as they need different techniques. The explanation of the cognitive parts is straightforward. Due to their symbolic nature, they are inherently interpretable so that goals, beliefs, and intentions can often simply be externalized by e.g. verbal communication. On the other hand, the post hoc explanation of conventional DNNs is much less straightforward, with a diverse research landscape.

The majority of explanation techniques for deep neural networks try to find and visualize relevant features in the input space that are responsible for classification \cite{montavon2018methods,samek2019towardsXAI,lapuschkin2019unmasking}.
The approaches can be divided into black-box methods that do not assume specific underlying structures and are only concerned with input-output pairings, e.g. local interpretable model-agnostic explanations (LIME) or meaningful perturbations \cite{ribeiro2016lime,fong2017interpretable}, and methods that take into account the distributed hierarchical architecture and propagation mechanism of neural networks, like Grad-CAM or layerwise relevance propagation (LRP) \cite{selvaraju2017gradcam,montavon2019lrp}.
Grad-CAM, for example, analyses the gradients in the classification layer of convolutional neural networks and uses the gradient flow back into preceding layers to construct a heatmap of important regions in the input space. By overlaying the heatmap with an input image for a visual classification, relevant features of the image are highlighted. These input features can be considered domain-specific, low-level knowledge, and not necessarily interpretable because they can require substantial top-down knowledge and processing in order to be interpretable for the user. The amount of knowledge needed varies between modalities. Whereas words and images might be intuitively interpretable even for a lay user, other feature spaces are less intelligible to human thinking. A heatmap of specific frequencies in a spectrogram is beyond anyone's understanding, except perhaps, for experts in the respective field. 

Most works applying XAI techniques pertain to the visual field and image classifiers. Communicating visual explanations that highlight important features in the input image with the means of a humanoid robot is problematic for various reasons. The feature space of an input image is usually quite large and highly correlated. A single highlighted pixel value does not mean very much for a human interpreter whereas a neighborhood of pixels or several superpixels contains ambiguities. However, even if a subset of correlated pixels might be representing a concept, it must be identified by the explanation method as such. 

Explanations in an HRI context should, therefore, take into account that human minds do not operate in feature space. However, few works try to abstract from pure feature visualization to higher-level concepts which are more akin to human knowledge processing like concept-based reasoning \cite{kim2018interpretability}. However, even methods that can generate meaningful visual concepts for explanations stay on the sub-symbolic level and do not describe the concepts semantically \cite{ghorbani2019ace}. There is still a significant gap in the research literature of explaining neural networks with higher-level knowledge and abstracting from lower-level feature visualizations, especially outside the visual domain \cite{das2020opportunities}. Concept-level explanations are easier to transfer into semantic representations which would help robots communicate them.

However, it is not always straightforward to understand the explanation of a classification, even if it shows somewhat interpretable input features or even concepts.
Anchors, a promising explanation method, tries to find conjunctive rules that locally explain a classification \cite{ribeiro2018anchors}. In cases where the rules are chaining together inherently interpretable chunks like words, simple rules can explain decisions with a high degree of precision. Their limitation lies in not being able to capture the (non-linear) interaction of the features which might lead to false interpretations on the user's side. Imagine a situation where the user is presented with a seemingly linear explanation of a decision that is actually non-linear.
A further problem is that local explanations do not account for the global behavior of a model. An agent explaining a decision locally cannot give a guarantee of making a similar decision in a similar situation. This is problematic for HRI contexts, where the future prediction of robot behavior through explanations plays a big role in building a successful relationship between humans and robots \cite{klein2004ten}. Generalization from this rule without further information might prove catastrophic in predicting future agent behavior.

Hybrid agents can be regarded as being goal-driven and data-driven at the same time. While trying to achieve a certain objective, the perceptions they form about their environment shape the way they obtain objectives and ultimately, the realized behavior. By explaining both of these aspects separately, the task to make the connection and infer their interaction is left to the user which leaves a lot of room to ambiguities. 

The transfer of concepts and explanations from the sub-symbolic to the symbolic level is crucial when relaying information from robots to humans. After all, the explanations must be communicated within the means of the agent. Considering humanoid robots, except for special models like the Pepper \cite{pandey2018pepper}, they are confined to the channels of communication that humans are familiar with, i.e. speech and gestures. But relaying the importance of pixel values or regions in an image through speech alone is not very feasible. Explanations need therefore be explicitly catered to or transformed into a modality that is communicable by the robot.

\section{Getting NICO to explain}
\label{sec:nico}
\begin{figure}[t]
    \centering
    \includegraphics[width=0.9\columnwidth]{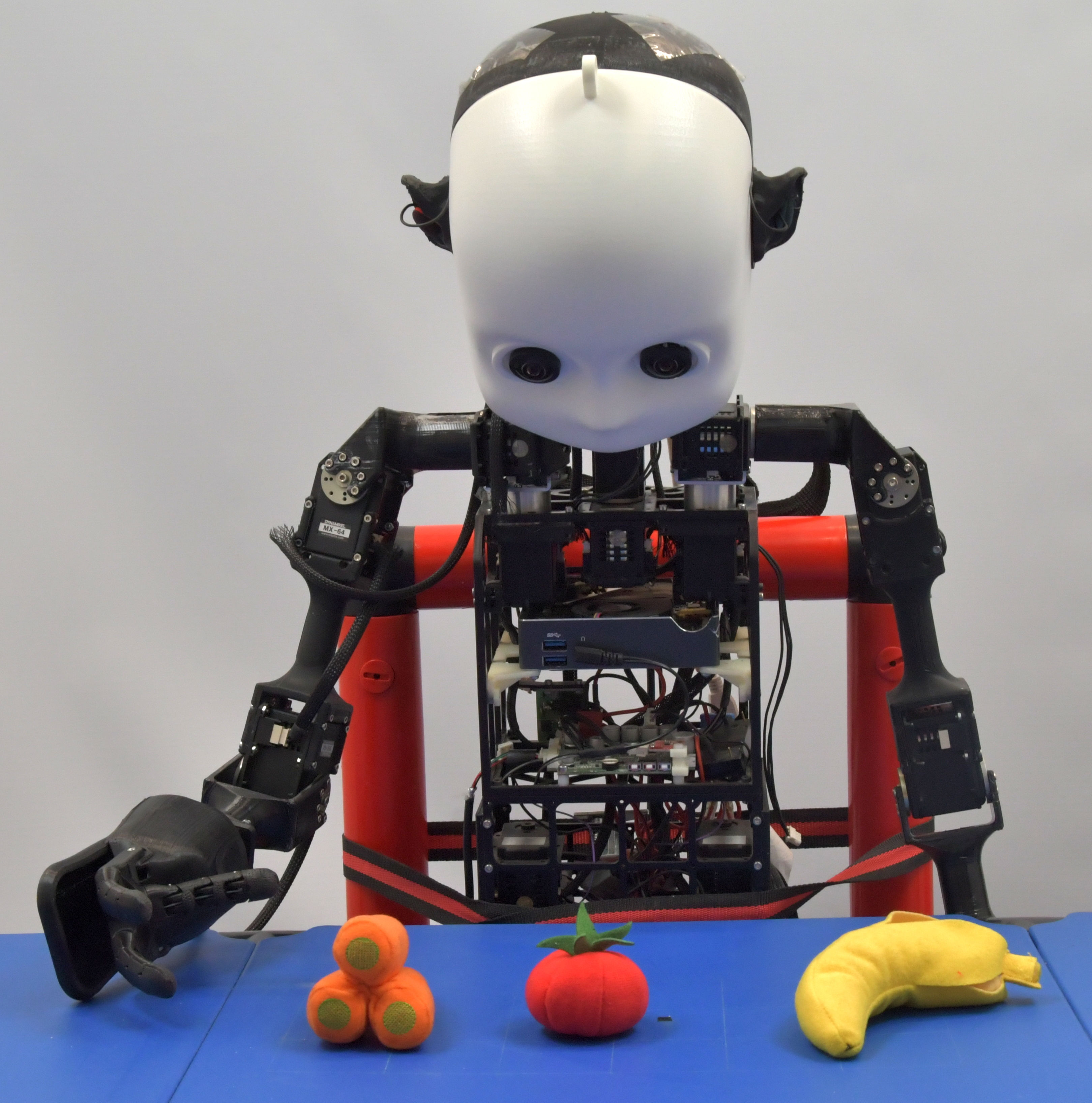} 
    \caption{The NICO is a multimodal, humanoid developmental robot built on an affordable and modular open-source framework for HRI research purposes.}
    \label{fig:nico}
\end{figure}

NICO, the Neuro-Inspired COmpanion, is a humanoid robot designed for human-robot interaction \cite{kerzel2017nico}. The NICO, as seen in figure \ref{fig:nico}, can capture data about its environment from two cameras as eyes for stereoscopic visual input, from two microphones as ears for auditive input, and from haptic sensors on his hands and fingers for tactile input. In order to communicate with humans, there are several LEDs in its head to simulate mouth and eyebrows to show facial expressions. Further, NICO possesses speakers to transmit sounds and speech. Lastly, two degrees of freedom (DoF) for yaw and pitch head movements and at least eight DoF per arm, depending on the exact hand model, allow to convey information via complex gestures. 
The platform that NICO is built upon mainly consists of dynamixel motors and open-sourced 3d printed parts. The robot can be flexibly adapted to specific needs depending on the experimental circumstances (for a more detailed description see \cite{kerzel2017nico}).

\begin{figure}[t]
    \centering
    \includegraphics[width=0.9\columnwidth]{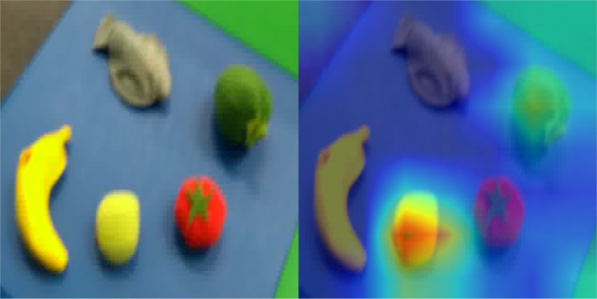}
    \caption{Left: NICO's visual input. Right: Heatmaps produced by NICO for the class "Apple" through Grad-CAM for weakly supervised object localization.}
    \label{fig:heatmap}
\end{figure}

Intrinsic explanations made by NICO would benefit the positive relations and successful teamwork with other collaborators. Understanding its own mechanisms can also help to gain useful insight and information about the environment, that was not explicitly learned. Suppose that NICO is given a command by a human collaborator and has the task to grasp one out of a few identifiable objects in front of it, but is not equipped with an explicit object localizer (for a visualization of the scene, see figure \ref{fig:nico}). An image classifier, like the Inception Network \cite{szegedy2015going}, only lets NICO identify the objects but not necessarily their location. Figure \ref{fig:nico} shows such a scenario. Once a specific object is identified, NICO can utilize a feature visualization technique like Grad-CAM to highlight the relevant parts of its visual input that are responsible for the classification. Figure \ref{fig:heatmap} shows the input image that NICO receives from one of its visual sensors on the left side and the application of the Grad-CAM algorithm for the classification of "apple" on the right side. By exposing those regions of the input image that contain the object in question, the explanation method Grad-CAM offers the possibility of weakly-supervised location. The heatmap can be converted into a binary mask by setting a threshold and, subsequently, a bounding box can be drawn around the mask \cite{selvaraju2017gradcam}. This way, NICO can learn to grasp objects without ever being exposed to localization training. Thus, intrinsic explanations can make an agent more versatile by exploiting the mechanisms of neural networks that were not explicitly trained.

Explanations should be explored in an interactive context with the goal of mutual understanding. The roles of explanator and explanee need not be fixed with robots being the former and humans being the latter \cite{ciatto2020agentbased}. Successful collaboration requires the robot to know about the human and vice versa. Representing each others' beliefs and cognitions in a Theory of Mind is necessary to predict the other's behavior, provide meaningful explanations, and, ultimately, trust each other \cite{shvo2020maayan,vinanzi2019trust}. Here, neural networks play a crucial role in enabling an agent with the ability for continual learning \cite{parisi2019continual}.
However, the use of deep learning models can result in learning unintended shortcuts that do not generalize well \cite{geirhos2020shortcut}. Intrinsic explanations made by NICO can help expose those shortcuts. Without the help of extrinsic explanations that describe better-suited features or a better classification process, those shortcomings cannot be overcome.

As laid out in the PeCoX framework, the mere generation of an explanation does not constitute the complete process of explaining. In order to effectively convey explanations, it is also necessary to communicate them properly, so that they can be successfully interpreted by the recipient \cite{neerincx2018pecox}. This holds especially true in HRI contexts where communication and interaction with a robot is the central idea.
Due to being designed as a humanoid companion, NICO underlies certain constraints in perceiving the environment. In fact, all the modalities that are accessible for sensing the environment, i.e. visual, auditive, and tactile, as well as all the modalities for communicating, i.e. gestures, facial expressions, and speech, are shared with humans. Hence, this constraint ensures that the modalities of the explanations made by NICO are inherently familiar and more interpretable to human information processing. 

The most suitable medium for exchanging explanations in HRI contexts are rules and relations. Not only can they be expressed verbally and carry clearly defined semantics, but they are also interpretable by humans and robots alike. 
Hybrid neural models like knowledge-based neural networks that combine symbolic knowledge with the learning capacity of neural networks \cite{towel1994knowledge,besold2017neuralsymbolic} are a promising method for interactive learning through mutual explanations. Robots can utilize the rules given by an explanator and insert their symbolic knowledge into their neural models to learn better behaviors by extrinsic explanations.

In order to illustrate the necessity of neuro-symbolic hybrid approaches for explainable human-robot interaction, once again picture the grasping example with NICO. One can illustrate the effectiveness of explanation methods by analyzing a failed interaction between NICO and a collaborator, for instance after NICO has grasped the wrong object specified by the human partner in the given command. Two main origins of failure can be indicated.
Either, NICO did not understand the verbal command correctly or it did not classify the correct object successfully. Whereas feature visualisation techniques like LRP allow NICO to identify the most salient words in the given command and to communicate them to the human partner, thus explaining what it understood to be important. 
However, the same cannot be said about the visual object classification component. After all, feature visualization as an explanation is only as intelligible as the features themselves. Semantic features like words are much easier to understand than the values of individual pixels. Also, it is unclear how a robot can effectively communicate such heatmaps to the interaction partner. Therefore, such feature importance methods are not suitable for modalities with features that are by themselves not meaningful for a human. This illustrates a common problem that given explanations from popular XAI algorithms do no translate well into human-machine interaction contexts and need adaptations and extensions based on symbolic approaches \cite{abdul2018trends}. 
For example, a knowledge-extraction approach that constructs a knowledge graph out of a deep neural network classification and extracts meaningful relations between features could be applied to NICO's visual input to explain the (mis)classification \cite{zhang2018interpreting}. Such relations are symbolically represented and can thus be communicated to the collaborator verbally.

\section{Conclusion}

Proficiently using neural networks and explaining their behavior necessitates that they are explainable in an intelligible manner by verbal communication. Only a few methods, so far, enable the representation of an explanation in concepts \cite{ribeiro2018anchors,kim2018interpretability,ghorbani2019ace}. Further research about how to extract rules and semantic knowledge from neural networks will prove worthwhile for explainable neurorobotics.

By communicating the \textit{intrinsic explanations} about their behavior, robots will be enabled to interact with collaborators to reach mutual understanding and expose flaws in reasoning and decision processes. Interaction partners can then provide \textit{extrinsic explanations} to correct their mistakes. Through inserting the corrected knowledge back into the neural modules, robots can achieve the possibility to learn from interactions in a meaningful manner.

\section{Acknowledgements}
The authors gratefully acknowledge partial support from the German Research Foundation DFG under project CML (TRR 169).

\bibliographystyle{aaai}
\bibliography{references}

\end{document}